%% file: main.tex
\documentclass[sigconf]{acmart}

\include{preamble}

\usepackage{makecell}
\usepackage{newfloat}
\usepackage{setspace}
\DeclareFloatingEnvironment[
   fileext=lol,
   name=Listing
]{listing}

\usepackage{cleveref}

\copyrightyear{2020}
\acmYear{2020}
\setcopyright{rightsretained}
\acmConference[ICAIF '20]{ACM International Conference on AI in Finance}{October 15--16, 2020}{New York, NY, USA}
\acmBooktitle{ACM International Conference on AI in Finance (ICAIF '20), October 15--16, 2020, New York, NY, USA}
\acmDOI{10.1145/3383455.3422567}
\acmISBN{978-1-4503-7584-9/20/10}

\begin{document}

\title{CryptoCredit: Securely Training Fair Models}

\author{Leo de Castro}
\affiliation{
  \institution{MIT CSAIL}
  \city{Cambridge}
  \state{Massachusetts}
}
\email{ldec@mit.edu}

\author{Jiahao Chen}
\orcid{0000-0002-4357-6574}
\affiliation{
  \institution{J.P.Morgan AI Research}
  \city{New York}
  \state{New York}
}
\email{jiahao.chen@jpmorgan.com}

\author{Antigoni Polychroniadou}
\affiliation{
  \institution{J.P.Morgan AI Research}
  \city{New York}
  \state{New York}
}
\email{antigoni.poly@jpmorgan.com}


\begin{abstract}

When developing models for regulated decision making,
sensitive features like age, race and gender cannot be used
and must be obscured from model developers to prevent bias.
However, the remaining features still need to be tested for correlation with
sensitive features, which can only be done with the knowledge of those features.
We resolve this dilemma using a fully homomorphic encryption scheme,
allowing model developers to train linear regression and logistic regression models
and test them for possible bias without ever revealing the sensitive features in the clear.
We demonstrate how it can be applied to leave-one-out regression testing,
and show using the adult income data set that our method is practical to run.

\end{abstract}




\settopmatter{printfolios=true} 

\maketitle

\section{Introduction}


    Major institutions are incorporating machine learning systems to drive decisions more efficiently and accurately in tasks such as loan approval, prediction of defaults and fraud detection. 
    While machine learning algorithms progress rapidly and transform the industry, they introduce the following challenge: \\

\emph{How can we ensure that machine learning algorithms do not inadvertently perpetuate bias that is inherent in the training data?}\\

In a well-known example of historical bias appearing in a machine learning model, in 2016 a model to predict future criminal recidivism turned out to be biased against African-Americans \citep{Angwin2016,Dressel2018}.

 More and more financial institutions have developed teams of machine learning experts to develop models of creditworthiness.
 These models are trained on historical credit data, which raises concerns that these models could perpetuate historical biases against marginalized groups \citep{Courchane2000,Barocas2016,Hurley2016}.
 To address this concern, some consumer financial institutions have also developed compliance teams that are responsible for ensuring that the methods to determine credit do not violate federal anti-discrimination laws.

 One of the basic distinctions between the compliance team and the machine learning team is that the compliance team is able to see all fields in the customer database, including those corresponding to classes protected by anti-discrimination law (i.e.\ age, race, sex, etc.) \citep{Chen2018}, while the machine learning team is only able to see a subset of the data fields. The compliance team determines which features the machine learning team is allowed to see and, subsequently, use to train their models. 

 Obviously, the fields directly corresponding to the protected classes are not allowed to be viewed by the machine learning team. However, biases in the data could allow a model to implicitly learn these protected fields. This is represented by previously unprotected fields having strong correlation with the protected field.
 A classic example of this phenomenon is race and zip code, known as redlining \citep{Cohen-Cole2011,Feldman2015}.
 Even if a machine learning model does not directly use the race field during training, if the zip code field is also used during training, then the model could implicitly learn the race field since, in many areas, race and zip codes are highly correlated. This would be a reason that the compliance team would remove the zip code field from the fields available to the machine learning team. 
    
However, discovering these correlations is often difficult, since the bulk of the machine learning expertise is concentrated in the machine learning teams, who are not allowed to see the protected fields. The machine learning team could potentially assess whether or not there is unconscious bias in the training data by building models to predict the prohibitive features from the rest of the features. This state of affairs introduces another challenge: \\

\emph{How can we ensure that privacy of prohibitive features is preserved when the model is trained?}\\

    To solve this seemingly paradoxical problem, we use cryptography.
    Modern cryptography employs a rigorous mathematical approach to compute on encrypted data. A cryptographic tool, called Fully Homomorphic Encryption (FHE) solves the problem of secure computation between two parties, Alice and Bob. More specifically, Alice has an input $x$ and Bob has a function $f$. Alice wishes to outsource the computation of $f(x)$ to Bob without revealing $x$. FHE allows Alice to encrypt her input $x$ into a ciphertext $\ct_x$ that can be safely sent to Bob. Bob can then homomorphically evaluate the function $f$ on the ciphertext $\ct_x$ to produce a new ciphertext $\ct_{f(x)}$ that encrypts $f(x)$. This resulting ciphertext $\ct_{f(x)}$ is returned to Alice, who decrypts to obtain $f(x)$.


\subsection{Our Contributions}

 
 
We frame the problem of securely developing a fair ML model as a secure computation outsourcing problem. More specifically, the compliance team has training data $X$ that must be kept private. At the same time, the compliance team has limited computing resources and needs to outsource the training of the ML model to the machine learning team. This problem can be solved with FHE, since the compliance team can encrypt the training data set $X$ and send only the ciphertext to the machine learning team. The machine learning team then trains their model over the encrypted data set, then returns the model to the compliance team.

Not all aspects of the training data set are hidden from the machine learning team. The feature labels (or \emph{schema}) are revealed to the machine learning team, and, in addition, the feature labels are indicated as either protected or unprotected. Only the unprotected features are used in training the ML models, but the protected features are very useful in evaluating fairness tests on these ML models.


In \Cref{sec:wald}, we introduce one of the simplest methods for fairness testing in regression models,
which are commonly used in econometrics and other financial applications \citep{Pope2011,Baines2014}.
The basic idea is leave-one-out regression, in which we train two regression models using the same feature set $X$, but including also $S$ in only one of the models.
The idea is then to test whether or not the coefficients in these two models show statistically significant differences.
We present the classical approach to hypothesis testing using Wald tests to test the equality of regression coefficients \citep{Serfling1980}.
In \Cref{section:system}, we show how to use our privacy preserving techniques to perform these LOO regression by training on the encrypted features. 
In \Cref{section:UCIAdult}, we apply our analysis of leave-one-out regression and Wald's test to the UCI Adult data set \cite{kohavi1996,UCIAdult}, a well-studied data set of extracted census data with many fields similar to a loan application. When training in a privacy preserving manner a linear model on this data set, we found that our test revealed biases in the training data that were learned by the model. Specifically, we found the the linear model learned the correlation between the ``Age'' feature and the ``Marital Status'' feature, with the ``Never Married'' feature and the ``Widowed'' feature having the strongest correlation. This aligns nicely with intuition.

In addition, we found that certain models were better than others at detecting biases in the training data. Specifically, while the linear regression model was able to detect strong biases in the training data, the logistic regression model did not seem to learn this correlation, despite having essentially the same accuracy as the linear model.

\subsection{Related Work}

\paragraph{Fairness in artificial intelligence}
In recent years, concerns over fairness in automated decision-making systems has received well-deserved attention \citep{Munoz2016,Whittaker2019},
particularly in predictive policing \citep{Angwin2016,Chouldechova2017,Berk2018,Dressel2018,Mehrabi2019}
and credit decisioning \citep{Courchane2000,Abuhamad2019,Chen2019}.
Fairness is enshrined in regulatory requirements for some applications, notably in credit \citep{Courchane2000,Chen2018}
and hiring decisions \citep{Conway1983,Feldman2015,Raghavan2019,Lambrecht2019}.
Several authors have criticized the use of automated decision making without regard to fairness or interpretability needs
in other applications \citep{Zarsky2016,CorbettDavies2017,Rudin2019,Selbst2019}.
However, most of the work focuses on binary outcome variables,
and less so on continuous outcomes which result from regression studies.
Nevertheless, the work of \citet{Kleinberg2016,Pleiss2017} on calibration notions of fairness deserve special attention,
as being applicable specifically to regression models with continuous outcomes.

As described in the introduction, our work is based on a direct analysis of the coefficients of regression models,
as is common in econometric analysis with a history going back at least to \citet{Conway1983}.
A similar (but not identical) treatment of linear regression models is given in \citet{Pope2011}.
Similar ideas also appear in the leave one covariate out (LOCO) method of \citet{Lei2017} for explaining variable-level importance.

\section{Background}
\label{sec:background}

    \subsection{Notation \& Terminology}
        \begin{definition}[Data Schema]
            For a data set $\calD$, denote the schema for $\calD$ as $\calS_\calD$. The schema of the dataset consists of the labels of the features of the dataset. 
        \end{definition}

        \begin{definition}[Data Set Select Operation]
            Let $\calD$ be a data set with schema $\calS_\calD$, and let $\calS_\calT \subseteq \calS_\calD$ denote a subset of the features in $\calS_\calD$. The $\select$ operation takes in a subset of features $\calS_\calT$ along with a data set $\calD$ and produces a new dataset $\calT$ with the same entries as in $\calD$ except with only the features in $\calS_\calT$.
            $$ \calT \gets \select(\calS_\calT, \calD) $$
        \end{definition}

        For a subset $T$ of a set $S$, denote the compliment of $T$ in $S$ by $T^c$, where $T^c \cap T = \emptyset$ and $T^c \cup T = S$. We also denote $T^c = S \backslash T$.

    \subsection{Leave-One-Out Regression}

    In this section, we describe leave-one-out (LOO) regression,
    which is fundamental tool for the statistical tests we perform on machine learning models in this work. The goal of LOO regression is to determine how a given feature $f$ in a data set $\calD$ is related to the other features in the data set. The high-level idea is to train a model $\calM$ on the data set $\calD$, then train a second model $\calM'$ on the data set $\calD \backslash \{f\}$, then measure how the coefficients of the model $\calM'$ differ from the original model $\calM$. If the weights corresponding to another feature $f'$ change significantly, then we conclude that $f'$ and $f$ are correlated.

    \subsection{Cryptographic Notions}
    In this section, we give high level definitions for the cryptographic primitives that we will use in the protocols below. All of these primitives will be used in a black-box manner.

    A fully homomorphic encryption scheme allows us to perform arbitrary computations on encrypted data without decrypting it. Although the notion was introduced back in \citep{RivestAD1978}, the first implementation is due to a much later breakthrough of \citep{Gentry09}, and has now been followed with much exciting activity \citep{DijkGHV10,BrakerskiV11,BrakerskiGV12}. In a nutshell, the data owner can encrypt a message $m$ using a private key to produce some ciphertext $c$, which can be sent to some external party. The external party can then compute any function $f$ on the ciphertext, which will result in a new ciphertext $c'$ that corresponds to $f(m)$. The data owner can then decrypt $c'$ to obtain $f(m)$.

    FHE solves the problem of secure computation between two parties, a compliance team \Compteam and a machine learning team \MLteam: \Compteam encrypts their input and sends it to \MLteam, who can homomorphically evaluate the function on \Compteam's encrypted input and its own input, sending only the resulting ciphertext back to \Compteam, who can decrypt. 
    
    
    

    We formalize the algorithms that comprise a fully homomorphic encryption scheme.
    \begin{definition}[Fully Homomorphic Encryption] \label{def:LeveledHE}
    A fully homomorphic encryption scheme $$\mathsf{FHE} = (\KeyGen, \Encrypt, \Eval, \Decrypt)$$ is a set of probabilistic polynomial-time algorithms defined as follows:
    \begin{itemize}
        \item $\KeyGen(1^\lambda) \rightarrow (\sk, \pk, \evk)$\\
            Given the security parameter $\lambda$, outputs a key pair consisting of a public encryption key $\pk$, a secret decryption key $\sk$, and an evaluation key $\evk$.
        \item $\Encrypt(\pk, m) \rightarrow \ct$\\
            Given a message $m \in \mathcal{M}$ and an encryption key $\pk$, outputs a ciphertext $\ct$.
        \item $\Eval(\evk, f, \ct_1, \ct_2, \ldots, \ct_n) \rightarrow \ct'$\\
            Given the evaluation key, a description of a function $f\colon \mathcal{M}^n \rightarrow \mathcal{M}$, and $n$ ciphertexts encrypting messages $m_1, \ldots, m_n$, outputs the result ciphertext $\ct'$ encrypting $m' = f(m_1, \ldots, m_n)$.
        \item $\Decrypt(\sk, \ct) = m$\\
            Given the secret decryption key and a ciphertext $\ct$ encrypting $m$, outputs $m$.
    \end{itemize}
    \end{definition}


    We now briefly provide intuition for the security definition of an FHE scheme. Note that our only concern is security for the compliance team, and we will not consider any security notion for the machine learning team\footnote{This is often referred to as \emph{circuit privacy} or \emph{function privacy} in the FHE literature.}. The standard and most basic security definition of an encryption scheme is known as \emph{semantic security} \cite{GM82}. 
    
    \begin{definition}[Semantic Security \cite{GM82}] \label{def:semanticSecurity}
        An FHE scheme satisfied semantic security if no probabilistic polynomial time adversary $\calA$ can succeed in the following task with probability greater than $1/2 + negl(\lambda)$, where $negl(\lambda)$ is a function that shrinks faster than $1/poly(\lambda)$ for every polynomial $poly$.
        
        $\calA$ takes as input $\pk$ and $\evk$ and produces two messages $m_1, m_2 \in \calM$. $\calA$ then takes in the encryptions of $m_1$ and $m_2$, denoted $\ct_1$ and $\ct_2$, in a random order. $\calA$ must then decide which ciphertext is an encryption of $m_1$ and which ciphertext is an encryption of $m_2$.
    \end{definition}

    Intuitively, \Cref{def:semanticSecurity} says that the encryptions of two messages should be indistinguishable, even for adversarially chosen messages. As encryption schemes, all known FHE schemes satisfy this security definition.

\section{Leave-one-out regression and Wald's test}
\label{sec:wald}

In this section, we present a series of tests that we will perform on trained models to approach a model that is independent of a set of protected features.
The setting is formally similar to that of covariate adjustment \citep{Robinson1991},
and is also similar, but not identical to, the treatment of \citet{Pope2011}.
Let $S = \left(S_1, S_2, ..., S_{n_s}\right)$ be set of sensitive features,
$X = \left(X_1, X_2, ..., X_{n_x}\right)$ be the set of ordinary features,
and $Y$ be the target variable (with prediction $\hat Y$).

The idea is to create two linear regression models

\begin{align}
\calM : Y & = X \beta + S \gamma + \epsilon_\calM\textrm{, and} \label{eq:linearmodel1} \\
\calM': Y & = X \zeta + \epsilon_{\calM'}  \label{eq:linearmodel2},
\end{align}
where $\beta\in\mathbb R^{n_X}$, $\gamma\in\mathbb R^{n_S}$, and $\zeta\in\mathbb R^{n_X}$ are the 
vectors of regression coefficients and $\epsilon_\calM$, $\epsilon_{\calM'}$ are the corresponding noise terms.

Alternatively, we may also consider a similar pair of logistic regression models

\begin{align}
\calM : \log\frac{Y}{1-Y} & = X \beta + S \gamma + \epsilon_{\calM}\textrm{, and} \label{eq:logisticmodel1} \\
\calM': \log\frac{Y}{1-Y} & = X \zeta + \epsilon_{\calM'} \label{eq:logisticmodel2},
\end{align}
with similarly defined symbols.
To abuse notation, we distinguish linear regression from logistic regression modeling
by discussion context, keeping the same notation for consistency of discussion.
Nevertheless, $\beta_i$ will always refer to the regression coefficient for the $i$th feature,
regardless of regression model, and similarly for $\gamma_i$ and $\zeta_i$.

We can now use the generalized Wald's test to test constraints on the regression parameters \citep{Serfling1980}.
Let $N$ be number of data points, $n_X$ be the number of ordinary features, which is the number of columns of $X$,
$n_S$ be the number of sensitive features, which is the number of columns of $S$,
$D$ be a diagonal matrix with the first $n_X + n_S$ entries equal to $1/(N-n_X-n_S)$
and the next $n_X$ entries equal to $N - n_X$, whose $i$th diagonal entry
counts the degrees of freedom used to estimate the $i$th coefficient,
and $\theta = \begin{pmatrix}\beta & \gamma & \zeta\end{pmatrix}^T \in \mathbb R^{2n_X+n_S}$
be the vector of all coefficients belonging to both models.
Furthermore, let $\hat\theta = \begin{pmatrix}\hat\beta & \hat\gamma & \hat\zeta\end{pmatrix}^T \in \mathbb R^{2n_X+n_S}$
be the estimators for the parameters, and $\mathbb E(.)$ denote the expectation value.
To test hypotheses of the form $H \theta = c$, where $H$ is a matrix of size $k \times (2n_X+n_S)$,
construct the test statistic
\begin{equation}
W = \left(H \hat\theta - c\right)^T
    \left(H D^{1/2} \hat\Sigma D^{1/2} H^T \right)^{-1}
    \left(H \hat\theta - c\right)
    \sim \chi^2_k,
\end{equation}
where $\hat\theta$ are the estimates for the parameters,
and $\hat\Sigma = \mathbb E \left(
(\hat\theta - \mathbb E \hat\theta)
( \hat\theta - \mathbb E \hat \theta)^T
\right)$ is the estimated covariance matrix for the parameters.

This general form of Wald's test allows us to test multiple equalities simultaneously.
The multiple equality testing can be useful for testing hypotheses like $\beta = \zeta$,
testing equalities of all the coefficients simultaneously, $\gamma = 0$ for testing
for the null effect of $S$,
or more fine-grained tests where multiple dummy variables encoding the same categorical variable
can be tested.
Nevertheless, in this paper, we focus exclusively on hypothesis involving one equality test for simplicity,
while being illustrative of the general case.
More specifically, we test pairwise equality hypotheses of the form
$\beta_i = \zeta_i$,
each to be tested against their corresponding alternative hypothesis
$\beta_i \ne \zeta_i$.
Let $H_i$ be the $1 \times (2n_X + n_S)$ matrix with the $(1, i)$th entry equal to +1 and $(1, n_X + n_S + i)$th entry equal to -1,
and all other entries being 0.
The corresponding Wald statistic is then
\begin{equation}
\begin{aligned}
W = & \frac{\left(\hat\beta_i - \hat\zeta_i\right)^2}{
    \frac{\hat\Sigma_{\beta_i\beta_i}}{N-n_X-n_S} - \frac{\hat\Sigma_{\beta_i\zeta_i} + \hat\Sigma_{\zeta_i\beta_i}}{\sqrt{((N-n_X-n_S)(N-n_X)}}  + \frac{\hat\Sigma_{\zeta_i\zeta_i}}{N-n_X}
    } \\
  \approx & \frac{\left(\hat\beta_i - \hat\zeta_i\right)^2}{
    \se(\hat\beta_i)^2 + \se(\hat\zeta_i)
    } \sim \chi^2_{1}, 
\end{aligned}\label{eq:WaldTest}
\end{equation}
where we approximate $\hat\Sigma_{\beta_i\zeta_i} = \mathbb E \left((\hat\beta_i - \mathbb E \hat\beta_i)(\hat\zeta_i - \mathbb E \hat\zeta_i) \right) \approx 0$, and note that $\se(\hat\beta_i) = \sqrt{\Sigma_{\beta_i\beta_i}/(N-n_X-n_S)}$
and $\se(\hat\zeta_i) = \sqrt{\Sigma_{\zeta_i\zeta_i}/(N-n_X)}$ are exactly the definitions of the standard errors for the respective coefficients.

From \eqref{eq:WaldTest}, we have \Cref{algo:WaldTest}, which allows us to determine the coefficients in a model that are correlated with a protected feature.

\begin{algorithm}
\caption{Univariate Wald Test, denoted \waldtest}
\label{algo:WaldTest}
\begin{algorithmic}[1]
    \REQUIRE Two models $\calM$ and $\calM'$, and significance level $0<\alpha<1$.
    \STATE The models $\calM$ and $\calM'$ must satisfy the constraint that the training data $\calD$ for $\calM$ and training data $\calD'$ for model $\calM'$ satisfy the relation that $\calS_{\calD'} = \calS_\calD \cup \{p\} $, where $p$ is the protected feature.
    \STATE Compute the critical value $W^*$ from the $\chi^2_1$-distribution. For the standard $\alpha = 0.05$, $W^* = 3.9$.
    \STATE $\widetilde{S} \gets \emptyset$
    \FOR{each feature $i \in \calS_\calD$}
        \STATE Estimate the coefficients $\hat\beta_i$ and $\hat\zeta_i$ for the $i$th feature in models $\calM$
        and $\calM'$ respectively.
        \STATE Compute the standard errors $\se(\hat\beta_i)$ and $\se(\hat\zeta_i)$ of the coefficients $\hat\beta_i$ and $\hat\zeta_i$.
        \STATE Compute the Wald statistic, $W_i = \frac{\left(\hat\beta_i - \hat\zeta_i\right)^2}{\se(\hat\beta_i)^2 + \se(\hat\zeta_i)^2}$, of \eqref{eq:WaldTest}, testing the null hypothesis $\beta_i = \zeta_i$ against its alternative $\beta_i \ne \zeta_i$.
        \IF{$W_i > W^*$}
            \STATE $\widetilde{S} \gets \widetilde{S} \cup \{i\}$
        \ENDIF
    \ENDFOR
    \ENSURE $\widetilde{S}$
\end{algorithmic}
\end{algorithm}

\section{A Secure Fairness System}
\label{section:system}
    
    
    In this section, we present the design of our secure machine learning fairness protocol. The protocol is run between two parties: the machine learning party $\MLteam$ and the compliance party $\Compteam$.
    $\Compteam$ begins with a dataset $\calD = \calP \cup \calU$, where each entry in $\calD$ consists of a set of sensitive features $\calP$ and a set of unsensitive features $\calU$.
    We call features that are \emph{a priori} prohibited from use ``sensitive features", and features that are \emph{a priori} permitted to use ``unsensitive features". After the fairness tests are performed, we will have features in the unsensitive set that are correlated with the features in the sensitive set. None of these features may be used in the training data of the final model, and we call these features ``protected features". All features that are not protected are ``unprotected features." 
    
    The sets $\calP$ and $\calU$ are denoted by schema $\calS_\calP$ and $\calS_\calU$. At a high level, the goal of this protocol is to produce a model $\calM$ that is independent of all features in $\calP$. The approach is to identify features in $\calU$ that encode features in $\calP$. The protocol proceeds in a series of rounds, and at the end of each round the set $\calU$ is updated to exclude features correlated with features $\calP$. When $\calU$ is not updated at the end of the round, the model trained on the features in the most recent $\calU$ is accepted as a fair model. 
    
    The high level security goal of the protocol is to hide the values of the protected features from the machine learning team. By hiding these values, we can argue that these values could not have implicitly influenced their decision to include or exclude unprotected features. 
    
    \subsection{A Single Round of the Fairness Protocol}
    \label{subsection:singleRoundFairness}
    
    
    A single round of the protocol is given in \Cref{algo:ComplianceFairOneRound,algo:MLTeamFairOneRound}. The full protocol, presented in \Cref{algo:fullFairnessCompTeam,algo:fullFairnessMLTeam} in \Cref{subsection:fullFairnessProtocol}, primarily consists of running \Cref{algo:ComplianceFairOneRound,algo:MLTeamFairOneRound} repeatedly until the resulting model passes the Wald test. 
    
    Beginning with \Cref{algo:ComplianceFairOneRound}, the inputs to a single round for the compliance team is a data set $\calD$, a schema of sensitive features $\calS_\calP$, a schema of unprotected features $\calS_\calU$ and an $\FHE$ key pair $\pk$ and $\sk$. The schema of sensitive and unprotected features must partition the data set $\calD$. In other words, the two schema must satisfy $\emptyset = \calS_\calP \cap \calS_\calU$ and $\calS_\calD = \calS_\calP \cup \calS_\calU$, where $\calS_\calD$ is the schema for the whole data set. In each round of the protocol, the compliance team sends the encrypted dataset to the machine learning team along with the partition of sensitive and unprotected features, represented by $\calS_\calP$ and $\calS_\calU$. The machine learning team then trains a model on the unprotected features using a fully homomorphic encryption scheme. The result of this training is an encrypted model that the machine learning team returns to the compliance team. Since the compliance team has the $\FHE$ secret key $\sk$, they can decrypt the encrypted model to obtain the model weights in the clear. 
    
    In addition to the model $\calM$ trained on the unprotected features, the compliance team also receives a series of models $\calM'_i$ for each sensitive feature $p_i \in \calS_\calP$. The model $\calM'_i$ is trained on the dataset $\widetilde{\calU_i} = \select(\calS_\calU \cup \{p_i\}, \calD)$, which is all unprotected features plus the single sensitive feature $p_i$. For each of these models $\calM'_i$, the compliance team performs the univariate Wald test as described in \Cref{algo:WaldTest}. If there are features in $\calS_\calU$ that are correlated with the sensitive feature $p_i$ and the model $\calM$ has learned this correlation, the Wald test will detect this by examining the difference in the weights $\calM$ from the weights of $\calM'_i$. The output of this test will be the set of features $\widetilde{\calS_i} \subseteq \calS_\calU$ that are correlated with $p_i$. The features in $\widetilde{\calS_i}$ are now considered protected, and in subsequent rounds these features will be excluded from the set of unprotected features.    
    
    In \Cref{algo:MLTeamFairOneRound}, we have the machine learning team's view of a single round of the fairness protocol. The bulk of this algorithm is spent running the learning algorithm \Learn on the various encrypted data sets. The round for \MLteam begins by receiving the encrypted data set $\ct_\calD$ from the compliance team. \MLteam also receives a partition of the data set into sensitive and unprotected features, which are denoted by $\calS_\calP$ and $\calS_\calU$. \MLteam begins by performing a homomorphic \select operation in the encrypted data set $\ct_\calD$ to obtain an encryption of the data set with only the unprotected features, which we denote $\ct_\calU$. Next, \MLteam homomorphically evaluates their learning algorithm of choice using $\ct_\calU$ as input to produce an encryption of the resulting model, which we denote $\ct_\calM$.
    
    In order to produce the inputs for the univariate Wald fairness test, \MLteam must also train models for each of the sensitive features. For each feature $p_i \in \calS_\calP$, \MLteam produces a new set of training data $\widetilde{\calU_i} = \select(\calS_\calU \cup \{p_i\}, \calD)$. This operation must be performed homomorphically over the encrypted data set $\ct_\calD$ to obtain $\ct_{\widetilde{\calU_i}}$, an encryption of $\widetilde{\calU_i}$. Using this encrypted training data, \MLteam again runs the learning algorithm to obtain an encryption of the model $\calM_{p_i}$. The encryptions of these models $\ct_{\calM_{p_i}}$ along with the encryption of the real model $\ct_\calM$ are returned to the compliance team. The compliance team then performs fairness tests, and the results of the fairness tests are represented in the inputs to the next round.

    \begin{algorithm}
    \caption{Single Round of Fairness Protocol for $\Compteam$, denoted $\CompOneRound$}
    \label{algo:ComplianceFairOneRound}
    \begin{algorithmic}[1]
    \REQUIRE Data set $\calD$ \\ 
    \inputListIndent Schema of sensitive features $\calS_\calP$ \\
    \inputListIndent Scheme of unprotected features $\calS_\calU$ \\
    \inputListIndent $\FHE$ public key $\pk$\\
    \inputListIndent $\FHE$ secret key $\sk$.

    \STATE Encrypt the dataset $\ct_\calD \gets \Encrypt(\pk, \calD)$. 
    \STATE Send the encrypted dataset $\ct_\calD$ along with the two schemas $\calS_\calP$ and $\calS_\calU$ to \MLteam.
    \STATE Receive encrypted model $\ct_\calM$ trained on $\calS_\calU$ from \MLteam.
    \STATE $\calM \gets \FHE.\Decrypt(\sk, \ct_\calM)$
    \STATE $\widetilde{\calS} \gets \emptyset$
    \FOR{$p_i \in \calS_\calP$} 
        \STATE Receive encrypted model $\ct_{\calM_i'}$ from $\MLteam$.
        \STATE $\calM_{p_i} \gets \FHE.\Decrypt(\sk, \ct_{\calM_i'})$
        \STATE $\widetilde{\calS}_{i} \gets \waldtest(\calM,\calM_i')$ 
        \STATE $\widetilde{\calS} \gets \widetilde{\calS_i} \cup \widetilde{\calS}$
    \ENDFOR
    \ENSURE $\calM, \widetilde{\calS}$ 
    \end{algorithmic}
    \end{algorithm}
    
    \begin{algorithm}
    \caption{Single Round of the Fairness Protocol for $\MLteam$, denoted $\MLOneRound$}
    \label{algo:MLTeamFairOneRound}
    \begin{algorithmic}[1]
    \REQUIRE $\FHE$ evaluation key $\evk$\\
    \inputListIndent Learning algorithm $\Learn$.
    \STATE Receive the encrypted dataset $\ct_\calD$. 
    \STATE Receive the schema for the sensitive and unprotected features $\calS_\calP$ and $\calS_\calU$.
    \STATE Select the features for the training data $\calS_\calT \subseteq \calS_\calU$.
    \STATE $\ct_\calU \gets \Eval(\evk, \select(\calS_\calU,\  \cdot \ ),\ \  \ct_\calD)$.
    \STATE $\ct_\calM \gets \Eval(\evk, \Learn(\cdot), \ \ct_\calU)$
    \STATE Send $\ct_\calM$ to $\Compteam$.
    \FOR{$p_i \in \calS_\calP$} 
        \STATE $\calS_{\widetilde{\calU_i}} \gets \calS_\calU \cup \{p_i\}$
        \STATE $\ct_{\widetilde{\calU_i}} \gets \Eval(\evk, \select(\calS_{\widetilde{\calU_i}}, \ \cdot \ ), \ \ \ct_\calD)$
        \STATE $\ct_{\calM_{p_i}} \gets \Eval(\Learn(\cdot),\ \ \ct_{\widetilde{\calU_i}})$
        \STATE Send $\ct_{\calM_{p_i}}$ to $\Compteam$.
    \ENDFOR 
    \end{algorithmic}
    \end{algorithm}

    \subsection{Full Fairness Protocol}
    \label{subsection:fullFairnessProtocol}
    
    
    Building off of the algorithms in \Cref{subsection:singleRoundFairness}, we can define our full fairness protocol in \Cref{algo:fullFairnessCompTeam,algo:fullFairnessMLTeam}. These algorithms follow naturally from the discussion in \Cref{subsection:singleRoundFairness}, as they are essentially just running the single round algorithms until the machine learning team produces a model that is sufficiently independent of the sensitive features. 
    
    At the end of each round, the compliance team obtains a candidate model and a set of protected features $\widetilde{S}$ in the candidate model's training data that are correlated with the sensitive features. In the next round, the compliance team removes the features in $\widetilde{S}$ from the unprotected features $\calS_\calU$, then uses this updated $\calS_\calU$ in the input for the next round. If $\widetilde{S} = \emptyset$, then the candidate model is accepted as a fair model. 
    
    \begin{algorithm}
    \caption{Full Fairness Protocol for $\Compteam$}
    \label{algo:fullFairnessCompTeam}
    \begin{algorithmic}[1]
        \REQUIRE Data set $\calD$ \\
        \inputListIndent Schema of protected features $\calS_\calP$ \\ 
        \inputListIndent Security parameter $\lambda$.
        \STATE $(\sk, \pk, \evk) \gets \FHE.\KeyGen(1^\lambda)$
        \STATE Send $\evk$ to $\MLteam$.
        \WHILE{$\mathsf{True}$}
            \STATE Let $\calS_\calU = \calS_\calP^c$
            \STATE $\calM, \widetilde{\calS} \gets \CompOneRound(\calD, \calS_\calP, \calS_\calU, \pk, \sk) $
            \IF{$\widetilde{\calS}$ equals $\emptyset$}
                \STATE Send \texttt{Terminate} signal to $\MLteam$.
                \STATE \textbf{Return:} $\calM$
            \ENDIF
            \STATE Send \texttt{Continue} signal to $\MLteam$.
            \STATE $\calS_\calU = \calS_\calU \backslash \widetilde{\calS}$
        \ENDWHILE
    \end{algorithmic}
    \end{algorithm}
    
    \begin{algorithm}
    \caption{Full Fairness Protocol for $\MLteam$}
    \label{algo:fullFairnessMLTeam}
    \begin{algorithmic}[1]
        \REQUIRE Learning algorithm $\Learn$.
        \STATE Receive $\FHE$ evaluation key $\evk$ from $\Compteam$.
        \WHILE{$\mathsf{True}$}
            \STATE Run $\MLOneRound(\evk, \Learn)$
            \STATE Receive \texttt{Signal} from $\Compteam$.
            \IF{\texttt{Signal} equals \texttt{Terminate}}
                \STATE \textbf{Return} 
            \ENDIF
        \ENDWHILE
    \end{algorithmic}
    \end{algorithm}
    
    We can see that \Cref{algo:fullFairnessCompTeam} always terminates, since $\widetilde{\calS} \subseteq \calS_\calU$, so $\calS_\calU$ must shrink with each iteration of the \texttt{while} loop. If $\calS_\calU$ does not shrink, then it must be that $\widetilde{S} = \emptyset$, so the \texttt{while} loop terminates. 
    Similarly, we have that \Cref{algo:fullFairnessMLTeam} terminates exactly when  \Cref{algo:fullFairnessCompTeam} terminates. Since we have already argued that  \Cref{algo:fullFairnessCompTeam} terminates, we have that \Cref{algo:fullFairnessMLTeam} terminates as well.

    \subsection{Security}
    
    We now briefly and informally argue the security of the fairness protocol defined by \Cref{algo:fullFairnessCompTeam,algo:fullFairnessMLTeam}. Recall that our goal is to hide the values of the protected features from the machine learning team. Since we do not know which features will be in the protected set at the start of the protocol, we have the protocol hide all features from the machine learning team. 
    
    The remainder of the security argument follows directly from the semantic security of the $\FHE$ scheme, since the only information sent to the machine learning team related to the values of the data are $\FHE$ ciphertexts, which hide the encrypted values. 

\section{Case Study: UCI Adult Dataset} \label{section:UCIAdult}

In this section, we implement the system described in \Cref{section:system} to train a linear model on the adult income dataset,
a popular data set which is available at the machine learning dataset repository of the University of California at Irvine \citep{kohavi1996,UCIAdult}.
The UCI Adult data set contains 14 attributes extracted from 1994 census data,
of which 3 are federally protected classes: age, sex and race.
For simplicity, we consider age only in our study.
For the remaining 11 features,
we use a standard dummy variable transformation of the categorical variables,
also known as one-hot encoding,
encoding a categorical variable with $m$ levels into $m-1$ binary variables,
dropping one of the dummy variables to avoid collinearity issues in the subsequent regression.
This transformation of the categorical variables yielded a total of $n_X = 87$ features, being a mix of numeric and binary variables.
Dropping all rows with missing entries yields a final data set of $30,162$ rows in the training subset
and $15,060$ rows in the testing subset, which we use in our analysis.
The target feature is a binary variable indicating if the person corresponding to the entry has an income of at least \$50,000 per year.
Denote this set of $n_X=87$ features by \Adult and the age protected class as \Age, represented by a single positive integer.
We begin our experiments with $\calU = \Adult$ and $\calP = \{\Age\}$. 

\subsection{Linear Regression Model}

We began by training two models using the linear regression learning algorithm. The first model $\calM$ was trained on $\calU$ and achieves an accuracy of 83\%\footnote{The trivial accuracy for the UCI Adult data set is 75.4\%, which is the proportion of persons in the data set making less than \$50,000 dollars per year.}. The second model $\calM'$ was trained on the dataset $\calU \cup \{\Age\}$ and also achieved an accuracy of 83\%. The train-test split of the data entries were identical for both models. 

We then compute the \waldtest on inputs $\calM$ and $\calM'$ to obtain the set of features that the model $\calM$ correlates with \textsf{Age},
and calculate the corresponding $p$-values to test at the usual $\alpha=0.05$ level of significance.
We found two features that changed significantly out of the $n_X = 87$ coefficients, and these results are shown in full in \Cref{tab:linear},
with the significant features summarized in \Cref{table:firstMaritalStatusBias}.
These features are the Never-married and Widowed categories of marital status.
The significance of the Never-married category persists even after applying the very conservative Bonferroni correction for
multiple hypothesis testing \citep{bonferroni}, at the $\alpha/n_X = 0.0006$ level.


\begin{table}
    \centering
    \begin{tabular}{|c|c|c|c|c|}\hline
    & \makecell{Original\\Coefficient} & \makecell{Coefficient\\with Age} & \makecell{$W$} & $p$ \\ \hline
    Never Married  & -0.0775  & -0.0063 & 12.041 & 0.0005 \\ \hline
    Widowed  &  0.0895 & 0.009 & 4.445 & 0.0350 \\ \hline
    \end{tabular}
    \caption{Features with the largest difference in standard errors between the OLS model without Age and the OLS model with Age. These were the only features that changed more than two standard errors.}
    \label{table:firstMaritalStatusBias}
\end{table}

To account for this correlation in the marital status feature, we reduce the marital status feature from seven categories to a single binary category of \{Unmarried, Married\} in the next round. We repeat this experiment with the unprotected features updated to reflect this change to the marital status. Both of the resulting models remained around 83\% accuracy and none of the features between the two models changed significantly. We can then conclude that this final model trained on the unprotected \Adult data set with the updated marital status feature is independent of the protected \Age feature.

\subsection{Logistic Regression Model}

While biases may be present in the \Adult data set, not all models will learn this bias equally. To illustrate this, we performed the leave-one-out regression experiment with $\calU = \Adult$ and $\calP = \{\Age\}$ using a logistic regression learning model. Once again, the two models achieved 83\% accuracy, but, unlike the linear regression model, the weights in the logistic regression model did not change significantly when the \Age feature was added. Based on this test, we can conclude that this logistic regression model has not detected the biases in the \Adult data set that the linear regression model was able to learn. It is an interesting question for future work to consider how different models learn biases. The full results of this experiment are shown in \Cref{tab:logistic}.

\subsection{Encrypted Implementation}

The protocol in \Cref{section:system} is agnostic to the underlying homomorphic encryption scheme as well as the method in which the learning algorithm is implemented using the chosen homomorphic encryption scheme. Implementations of logistic regression using homomorphic encryption abound \citep{CKKS_SecureLogReg, BHHH19}. We implemented secure logistic regression training on the $\Adult$ data set using the HELR library from \citep{CKKS_SecureLogReg} on a standard machine\footnote{We used an AWS instance with a 3.0 GHz CPU with 8 cores and 32 GB of RAM.}. While the full running time for training a single logistic regression model was about 45 minutes, only about three minutes of computation was actually spent progressing the logistic regression function. The remainder of the computation time was spent on the bootstrapping operation. This operation does not advance the underlying computation; rather, it is a technical operation necessary to maintain the correctness of the result when the function circuit is sufficiently deep. However, our goal in this setting is not to hide the learning algorithm from the compliance team, only to minimize the amount of machine learning expertise on the compliance side. This allows us to employ much more efficient \emph{interactive} bootstrapping routine that can be run by the compliance team. By introducing interaction, we can effectively eliminate almost all of the computation time spent on the bootstrapping operation. It is an interesting direction for future work to optimize this fairness outsourcing protocol.

\begin{table}
\centering
\resizebox{0.75\columnwidth}{!}{%
\begin{tabular}{r|c|c|c|c|c|c}
Feature                    & $\hat\beta_i$ & $\se(\hat\beta_i)$ & $\hat\zeta_i$ & $\se(\hat\zeta_i)$ & $W_i$ & $p$ \\
\hline
Local-gov                  & -0.2172     & 0.0280         & -0.2132     & 0.0280         & 0.0102      & 0.9195  \\
Private                    & -0.1558     & 0.0240         & -0.1354     & 0.0240         & 0.3613      & 0.5478  \\
Self-emp-inc               & 0.0344      & 0.0320         & 0.0234      & 0.0320         & 0.0591      & 0.8080  \\
Self-emp-not-inc           & -0.2514     & 0.0280         & -0.2591     & 0.0280         & 0.0378      & 0.8458  \\
State-gov                  & -0.2611     & 0.0310         & -0.2469     & 0.0300         & 0.1084      & 0.7420  \\
Without-pay                & -0.4112     & 0.1900         & -0.4560     & 0.1890         & 0.0279      & 0.8672  \\
\hline
11th                       & 0.1069      & 0.0470         & 0.1826      & 0.0470         & 1.2971      & 0.2547  \\
12th                       & 0.2195      & 0.0770         & 0.3491      & 0.0770         & 1.4164      & 0.2340  \\
1st-4th                    & -0.2989     & 0.1320         & -0.5680     & 0.1330         & 2.0623      & 0.1510  \\
5th-6th                    & -0.2225     & 0.1000         & -0.4277     & 0.1000         & 2.1054      & 0.1468  \\
7th-8th                    & -0.2517     & 0.0680         & -0.4027     & 0.0680         & 2.4655      & 0.1164  \\
9th                        & -0.1275     & 0.0490         & -0.1974     & 0.0490         & 1.0175      & 0.3131  \\
Assoc-acdm                 & 0.5862      & 0.1850         & 0.9424      & 0.1860         & 1.8436      & 0.1745  \\
Assoc-voc                  & 0.5150      & 0.1560         & 0.8106      & 0.1560         & 1.7953      & 0.1803  \\
Bachelors                  & 0.8746      & 0.2140         & 1.2797      & 0.2150         & 1.7834      & 0.1817  \\
Doctorate                  & 1.5701      & 0.3050         & 2.1082      & 0.3060         & 1.5512      & 0.2130  \\
HS-grad                    & 0.2703      & 0.0960         & 0.4459      & 0.0970         & 1.6556      & 0.1982  \\
Masters                    & 1.1515      & 0.2440         & 1.5947      & 0.2450         & 1.6429      & 0.1999  \\
Preschool                  & -0.2939     & 0.1790         & -0.6219     & 0.1800         & 1.6695      & 0.1963  \\
Prof-school                & 1.4820      & 0.2750         & 1.9854      & 0.2760         & 1.6694      & 0.1963  \\
Some-college               & 0.4193      & 0.1260         & 0.6595      & 0.1260         & 1.8171      & 0.1777  \\
\hline
education                  & -0.0646     & 0.0300         & -0.1214     & 0.0300         & 1.7924      & 0.1806  \\
\hline
Married-AF-spouse          & 0.3060      & 0.1610         & 0.3992      & 0.1610         & 0.1676      & 0.6823  \\
Married-civ-spouse         & 0.1936      & 0.0520         & 0.2313      & 0.0510         & 0.2679      & 0.6047  \\
Married-spouse-absent      & 0.0558      & 0.0390         & 0.0687      & 0.0390         & 0.0547      & 0.8151  \\
\textbf{Never-married}     & -0.0775     & 0.0140         & -0.0063     & 0.0150         & \textbf{12.0414}     & \textbf{0.0005}**  \\
Separated                  & 0.0117      & 0.0260         & 0.0289      & 0.0250         & 0.2274      & 0.6335  \\
\textbf{Widowed}                    & 0.0895      & 0.0270         & 0.0090      & 0.0270         & \textbf{4.4446}      & \textbf{0.0350}  \\
\hline
Armed-Forces               & -0.2248     & 0.2360         & -0.1841     & 0.2350         & 0.0149      & 0.9027  \\
Craft-repair               & -0.0058     & 0.0170         & 0.0022      & 0.0170         & 0.1107      & 0.7393  \\
Exec-managerial            & 0.2934      & 0.0170         & 0.2877      & 0.0170         & 0.0562      & 0.8126  \\
Farming-fishing            & -0.1858     & 0.0270         & -0.1886     & 0.0270         & 0.0054      & 0.9415  \\
Handlers-cleaners          & -0.0847     & 0.0230         & -0.0673     & 0.0230         & 0.2862      & 0.5927  \\
Machine-op-inspct          & -0.0845     & 0.0200         & -0.0771     & 0.0200         & 0.0685      & 0.7936  \\
Other-service              & -0.0476     & 0.0170         & -0.0399     & 0.0170         & 0.1026      & 0.7488  \\
Priv-house-serv            & 0.0175      & 0.0610         & -0.0027     & 0.0610         & 0.0548      & 0.8149  \\
Prof-specialty             & 0.1426      & 0.0180         & 0.1457      & 0.0180         & 0.0148      & 0.9031  \\
Protective-serv            & 0.1785      & 0.0310         & 0.1871      & 0.0310         & 0.0385      & 0.8445  \\
Sales                      & 0.1013      & 0.0170         & 0.1052      & 0.0170         & 0.0263      & 0.8711  \\
Tech-support               & 0.1557      & 0.0260         & 0.1616      & 0.0260         & 0.0257      & 0.8725  \\
Transport-moving           & -0.0492     & 0.0220         & -0.0491     & 0.0220         & 0.0000      & 0.9974  \\
\hline
Not-in-family              & -0.4286     & 0.0510         & -0.3988     & 0.0510         & 0.1707      & 0.6795  \\
Other-relative             & -0.3836     & 0.0500         & -0.3381     & 0.0500         & 0.4141      & 0.5199  \\
Own-child                  & -0.4229     & 0.0510         & -0.3453     & 0.0510         & 1.1576      & 0.2820  \\
Unmarried                  & -0.4503     & 0.0530         & -0.4094     & 0.0520         & 0.3034      & 0.5817  \\
Wife                       & 0.1134      & 0.0200         & 0.1354      & 0.0200         & 0.6050      & 0.4367  \\
\hline
hours-per-week             & 0.0061      & 0.0000         & 0.0065      & 0.0000         & N/A         & N/A     \\
\hline
Canada                     & -0.2482     & 0.1800         & -0.2576     & 0.1790         & 0.0014      & 0.9705  \\
China                      & -0.5128     & 0.1870         & -0.5120     & 0.1860         & 0.0000      & 0.9976  \\
Columbia                   & -0.5449     & 0.1910         & -0.5484     & 0.1900         & 0.0002      & 0.9896  \\
Cuba                       & -0.3161     & 0.1820         & -0.3392     & 0.1810         & 0.0081      & 0.9283  \\
Dominican-Republic         & -0.4350     & 0.1870         & -0.4316     & 0.1860         & 0.0002      & 0.9897  \\
Ecuador                    & -0.4091     & 0.2140         & -0.3985     & 0.2140         & 0.0012      & 0.9721  \\
El-Salvador                & -0.3361     & 0.1810         & -0.3129     & 0.1800         & 0.0083      & 0.9276  \\
England                    & -0.2557     & 0.1830         & -0.2604     & 0.1820         & 0.0003      & 0.9855  \\
France                     & -0.2318     & 0.2150         & -0.2215     & 0.2140         & 0.0012      & 0.9729  \\
Germany                    & -0.2554     & 0.1780         & -0.2500     & 0.1770         & 0.0005      & 0.9828  \\
Greece                     & -0.4793     & 0.2120         & -0.5059     & 0.2110         & 0.0079      & 0.9291  \\
Guatemala                  & -0.2932     & 0.1890         & -0.2505     & 0.1880         & 0.0257      & 0.8727  \\
Haiti                      & -0.3744     & 0.1990         & -0.3811     & 0.1980         & 0.0006      & 0.9810  \\
Holand-Netherlands         & -0.3151     & 0.7240         & -0.3421     & 0.7210         & 0.0007      & 0.9789  \\
Honduras                   & -0.3570     & 0.2630         & -0.3290     & 0.2620         & 0.0057      & 0.9399  \\
Hong                       & -0.4256     & 0.2320         & -0.3862     & 0.2310         & 0.0145      & 0.9042  \\
Hungary                    & -0.3617     & 0.2570         & -0.4017     & 0.2560         & 0.0122      & 0.9122  \\
India                      & -0.4434     & 0.1810         & -0.4250     & 0.1800         & 0.0052      & 0.9425  \\
Iran                       & -0.3170     & 0.1990         & -0.3010     & 0.1980         & 0.0032      & 0.9545  \\
Ireland                    & -0.2232     & 0.2200         & -0.2232     & 0.2190         & 0.0000      & 1.0000  \\
Italy                      & -0.1962     & 0.1870         & -0.2148     & 0.1860         & 0.0050      & 0.9438  \\
Jamaica                    & -0.3773     & 0.1840         & -0.3750     & 0.1830         & 0.0001      & 0.9929  \\
Japan                      & -0.2389     & 0.1900         & -0.2299     & 0.1890         & 0.0011      & 0.9732  \\
Laos                       & -0.5419     & 0.2380         & -0.5245     & 0.2370         & 0.0027      & 0.9587  \\
Mexico                     & -0.4344     & 0.1690         & -0.3908     & 0.1680         & 0.0335      & 0.8548  \\
Nicaragua                  & -0.4587     & 0.2060         & -0.4318     & 0.2060         & 0.0085      & 0.9264  \\
Outlying-US(Guam-USVI-etc) & -0.6610     & 0.2510         & -0.6685     & 0.2500         & 0.0004      & 0.9831  \\
Peru                       & -0.4478     & 0.2100         & -0.4518     & 0.2090         & 0.0002      & 0.9892  \\
Philippines                & -0.2712     & 0.1740         & -0.2809     & 0.1730         & 0.0016      & 0.9685  \\
Poland                     & -0.4190     & 0.1910         & -0.4309     & 0.1900         & 0.0020      & 0.9648  \\
Portugal                   & -0.4434     & 0.2060         & -0.4196     & 0.2050         & 0.0067      & 0.9347  \\
Puerto-Rico                & -0.3881     & 0.1790         & -0.3922     & 0.1790         & 0.0003      & 0.9871  \\
Scotland                   & -0.5051     & 0.2700         & -0.5046     & 0.2690         & 0.0000      & 0.9990  \\
South                      & -0.5824     & 0.1860         & -0.5774     & 0.1850         & 0.0004      & 0.9848  \\
Taiwan                     & -0.4902     & 0.1990         & -0.4603     & 0.1980         & 0.0113      & 0.9152  \\
Thailand                   & -0.5519     & 0.2380         & -0.5231     & 0.2380         & 0.0073      & 0.9318  \\
Trinadad\&Tobago           & -0.4410     & 0.2350         & -0.4572     & 0.2340         & 0.0024      & 0.9610  \\
United-States              & -0.3001     & 0.1660         & -0.2983     & 0.1660         & 0.0001      & 0.9939  \\
Vietnam                    & -0.4805     & 0.1880         & -0.4686     & 0.1870         & 0.0020      & 0.9642  \\
Yugoslavia                 & -0.2402     & 0.2420         & -0.2316     & 0.2410         & 0.0006      & 0.9799 
\end{tabular}
}
\caption{Estimated coefficients $\beta$ and $\zeta$ in the respective linear regression models $\calM$ \eqref{eq:linearmodel1} and $\calM'$ \eqref{eq:linearmodel2}, along with their standard errors (se), Wald statistic \eqref{eq:WaldTest}, and its $p$-value
as determined from the $\chi^2_1$-distribution.
Bold $p$-values are significant at the $\alpha=0.05$ level; values marked ** are significant at the Bonferroni-corrected $\alpha/n_X = 0.0006$ level.
Only the Never-married and Widowed categories of marital status show statistically significant differences between $\beta_i$ and $\zeta_i$.}
\label{tab:linear}
\end{table}

\begin{table}
\resizebox{0.75\columnwidth}{!}{%
\begin{tabular}{r|c|c|c|c|c|c}
Feature                    & $\hat\beta_i$ & $\se(\hat\beta_i)$ & $\hat\zeta_i$ & $\se(\hat\zeta_i)$ & $W_i$ & $p$ \\
\hline
Local-gov                  & -1.2222 & 0.1050  & -1.1829     & 0.1040         & 0.0707 & 0.7903  \\
Private                    & -0.9785 & 0.0870  & -1.0847     & 0.0860         & 0.7537 & 0.3853  \\
Self-emp-inc               & -0.4591 & 0.1150  & -0.3583     & 0.1150         & 0.3841 & 0.5354  \\
Self-emp-not-inc           & -1.2819 & 0.1020  & -1.2774     & 0.1010         & 0.0010 & 0.9750  \\
State-gov                  & -1.4024 & 0.1170  & -1.3379     & 0.1150         & 0.1546 & 0.6942  \\
Without-pay                & -0.0642 & 0.7550  & -0.0580     & 0.7460         & 0.0000 & 0.9953  \\
\hline
11th                       & -1.2592 & 0.2220  & -1.2487     & 0.2200         & 0.0011 & 0.9732  \\
12th                       & -0.5439 & 0.3330  & -0.5270     & 0.3300         & 0.0013 & 0.9712  \\
1st-4th                    & -0.4908 & 0.5470  & -0.4476     & 0.5340         & 0.0032 & 0.9549  \\
5th-6th                    & -0.8337 & 0.4300  & -0.7778     & 0.4180         & 0.0087 & 0.9257  \\
7th-8th                    & -1.4437 & 0.2880  & -1.3697     & 0.2800         & 0.0339 & 0.8538  \\
9th                        & -1.0654 & 0.2330  & -1.0179     & 0.2260         & 0.0214 & 0.8837  \\
Assoc-acdm                 & -0.4001 & 0.7100  & -0.4239     & 0.7070         & 0.0006 & 0.9810  \\
Assoc-voc                  & -0.3023 & 0.5970  & -0.3400     & 0.5950         & 0.0020 & 0.9643  \\
Bachelors                  & 0.4935  & 0.8170  & 0.4311      & 0.8140         & 0.0029 & 0.9569  \\
Doctorate                  & 1.1556  & 1.1600  & 1.1284      & 1.1540         & 0.0003 & 0.9867  \\
HS-grad                    & -0.6676 & 0.3730  & -0.6800     & 0.3720         & 0.0006 & 0.9812  \\
Masters                    & 0.7324  & 0.9300  & 0.7498      & 0.9260         & 0.0002 & 0.9894  \\
Preschool                  & -0.1453 & 0.8130  & -0.1316     & 0.7860         & 0.0001 & 0.9903  \\
Prof-school                & 1.4555  & 1.0470  & 1.4311      & 1.0420         & 0.0003 & 0.9868  \\
Some-college               & -0.2981 & 0.4830  & -0.2713     & 0.4820         & 0.0015 & 0.9687  \\
\hline
education                  & 0.0066  & 0.1130  & -0.0096     & 0.1130         & 0.0103 & 0.9193  \\
\hline
Married-AF-spouse          & 0.0945  & 0.5810  & 0.0921      & 0.5870         & 0.0000 & 0.9977  \\
Married-civ-spouse         & 0.3700  & 0.2530  & 0.3461      & 0.2530         & 0.0045 & 0.9467  \\
Married-spouse-absent      & -0.3959 & 0.2140  & -0.3817     & 0.2060         & 0.0023 & 0.9619  \\
Never-married              & -1.0609 & 0.0760  & -1.0122     & 0.0760         & 0.2053 & 0.6505  \\
Separated                  & -0.5968 & 0.1560  & -0.6017     & 0.1520         & 0.0005 & 0.9821  \\
Widowed                    & -0.3541 & 0.1510  & -0.3160     & 0.1380         & 0.0347 & 0.8522  \\
\hline
Armed-Forces               & -0.0209 & 0.9310  & -0.0198     & 0.9240         & 0.0000 & 0.9993  \\
Craft-repair               & -0.1118 & 0.0720  & -0.0743     & 0.0710         & 0.1375 & 0.7107  \\
Exec-managerial            & 0.5813  & 0.0690  & 0.5542      & 0.0680         & 0.0783 & 0.7797  \\
Farming-fishing            & -1.1937 & 0.1310  & -1.1836     & 0.1280         & 0.0030 & 0.9560  \\
Handlers-cleaners          & -1.2222 & 0.1400  & -1.2311     & 0.1410         & 0.0020 & 0.9643  \\
Machine-op-inspct          & -0.8202 & 0.0970  & -0.8737     & 0.0980         & 0.1505 & 0.6980  \\
Other-service              & -1.6790 & 0.1250  & -1.7034     & 0.1240         & 0.0192 & 0.8898  \\
Priv-house-serv            & -0.2535 & 0.4060  & -0.2343     & 0.3790         & 0.0012 & 0.9724  \\
Prof-specialty             & 0.3759  & 0.0730  & 0.4542      & 0.0720         & 0.5832 & 0.4451  \\
Protective-serv            & 0.3820  & 0.1170  & 0.3813      & 0.1160         & 0.0000 & 0.9966  \\
Sales                      & 0.1872  & 0.0730  & 0.2486      & 0.0730         & 0.3537 & 0.5520  \\
Tech-support               & 0.6317  & 0.1010  & 0.6531      & 0.1000         & 0.0227 & 0.8803  \\
Transport-moving           & -0.3347 & 0.0910  & -0.3523     & 0.0900         & 0.0189 & 0.8906  \\
\hline
Not-in-family              & -1.2453 & 0.2510  & -1.1301     & 0.2520         & 0.1049 & 0.7460  \\
Other-relative             & -1.1723 & 0.2280  & -1.1185     & 0.2280         & 0.0278 & 0.8675  \\
Own-child                  & -2.1693 & 0.2570  & -2.1192     & 0.2600         & 0.0188 & 0.8910  \\
Unmarried                  & -1.9072 & 0.2630  & -1.8586     & 0.2630         & 0.0171 & 0.8960  \\
Wife                       & 0.2722  & 0.0680  & 0.2379      & 0.0680         & 0.1272 & 0.7213  \\
\hline
hours-per-week             & 0.0185  & 0.0020  & 0.0161      & 0.0020         & 0.7200 & 0.3961  \\
\hline
Canada                     & -0.0900 & 0.6610  & -0.0756     & 0.6500         & 0.0002 & 0.9876  \\
China                      & -0.2274 & 0.6900  & -0.2141     & 0.6800         & 0.0002 & 0.9890  \\
Columbia                   & -0.2045 & 0.7590  & -0.1919     & 0.7440         & 0.0001 & 0.9905  \\
Cuba                       & -0.1338 & 0.6770  & -0.1165     & 0.6640         & 0.0003 & 0.9854  \\
Dominican-Republic         & -0.2096 & 0.7630  & -0.1968     & 0.7450         & 0.0001 & 0.9904  \\
Ecuador                    & -0.0701 & 0.8420  & -0.0668     & 0.8290         & 0.0000 & 0.9978  \\
El-Salvador                & -0.2189 & 0.7160  & -0.2056     & 0.7020         & 0.0002 & 0.9894  \\
England                    & -0.0582 & 0.6750  & -0.0466     & 0.6630         & 0.0002 & 0.9902  \\
France                     & 0.0032  & 0.7930  & 0.0058      & 0.7760         & 0.0000 & 0.9981  \\
Germany                    & -0.1087 & 0.6510  & -0.0935     & 0.6400         & 0.0003 & 0.9867  \\
Greece                     & -0.0989 & 0.7740  & -0.0899     & 0.7660         & 0.0001 & 0.9934  \\
Guatemala                  & -0.1373 & 0.7910  & -0.1294     & 0.7740         & 0.0001 & 0.9943  \\
Haiti                      & -0.0921 & 0.8310  & -0.0854     & 0.8150         & 0.0000 & 0.9954  \\
Holand-Netherlands         & -0.0009 & 6.0310  & -0.0009     & 5.9830         & 0.0000 & 1.0000  \\
Honduras                   & -0.0231 & 1.4540  & -0.0219     & 1.3780         & 0.0000 & 0.9995  \\
Hong                       & -0.0479 & 0.8700  & -0.0453     & 0.8550         & 0.0000 & 0.9983  \\
Hungary                    & -0.0299 & 0.9050  & -0.0255     & 0.8830         & 0.0000 & 0.9972  \\
India                      & -0.2407 & 0.6610  & -0.2250     & 0.6520         & 0.0003 & 0.9865  \\
Iran                       & -0.0548 & 0.7270  & -0.0501     & 0.7150         & 0.0000 & 0.9963  \\
Ireland                    & -0.0119 & 0.8840  & -0.0105     & 0.8740         & 0.0000 & 0.9991  \\
Italy                      & -0.0552 & 0.6850  & -0.0428     & 0.6720         & 0.0002 & 0.9897  \\
Jamaica                    & -0.1542 & 0.7110  & -0.1436     & 0.7000         & 0.0001 & 0.9915  \\
Japan                      & -0.0446 & 0.7110  & -0.0400     & 0.6980         & 0.0000 & 0.9963  \\
Laos                       & -0.0587 & 0.9130  & -0.0554     & 0.9130         & 0.0000 & 0.9980  \\
Mexico                     & -1.4682 & 0.6420  & -1.4426     & 0.6300         & 0.0008 & 0.9773  \\
Nicaragua                  & -0.0993 & 0.8450  & -0.0945     & 0.8340         & 0.0000 & 0.9968  \\
Outlying-US(Guam-USVI-etc) & -0.0629 & 1.0150  & -0.0588     & 0.9810         & 0.0000 & 0.9977  \\
Peru                       & -0.0791 & 0.8690  & -0.0749     & 0.8550         & 0.0000 & 0.9973  \\
Philippines                & -0.1455 & 0.6420  & -0.1240     & 0.6310         & 0.0006 & 0.9809  \\
Poland                     & -0.1274 & 0.7090  & -0.1169     & 0.6990         & 0.0001 & 0.9916  \\
Portugal                   & -0.1234 & 0.7840  & -0.1158     & 0.7680         & 0.0000 & 0.9945  \\
Puerto-Rico                & -0.2873 & 0.6860  & -0.2685     & 0.6730         & 0.0004 & 0.9844  \\
Scotland                   & -0.0291 & 1.0110  & -0.0269     & 1.0010         & 0.0000 & 0.9988  \\
South                      & -0.2352 & 0.7020  & -0.2205     & 0.6920         & 0.0002 & 0.9881  \\
Taiwan                     & -0.0826 & 0.7430  & -0.0769     & 0.7350         & 0.0000 & 0.9956  \\
Thailand                   & -0.0486 & 0.9420  & -0.0460     & 0.9180         & 0.0000 & 0.9984  \\
Trinadad\&Tobago           & -0.0595 & 0.8860  & -0.0542     & 0.8680         & 0.0000 & 0.9966  \\
United-States              & -0.2211 & 0.6070  & -0.2960     & 0.5970         & 0.0077 & 0.9299  \\
Vietnam                    & -0.1795 & 0.7370  & -0.1692     & 0.7250         & 0.0001 & 0.9921  \\
Yugoslavia                 & -0.0097 & 0.9010  & -0.0086     & 0.8890         & 0.0000 & 0.9993 
\end{tabular}
}
\caption{
Estimated coefficients $\hat\beta$ and $\hat\zeta$ in the respective logistic regression models $\calM$ \eqref{eq:logisticmodel1} and $\calM'$ \eqref{eq:logisticmodel2}, with other notation as defined in \Cref{tab:linear}.
None of the features are found to show statistically significant differences between the coefficients $\beta_i$ and $\zeta_i$.
}
\label{tab:logistic}
\end{table}

{\small
\begin{spacing}{0.9}
\paragraph{Disclaimer}
This paper was prepared for informational purposes by the Artificial Intelligence Research group of JPMorgan Chase \& Co and its affiliates (``JP Morgan''), and is not a product of the Research Department of JP Morgan.  JP Morgan makes no representation and warranty whatsoever and disclaims all liability, for the completeness, accuracy or reliability of the information contained herein.  This document is not intended as investment research or investment advice, or a recommendation, offer or solicitation for the purchase or sale of any security, financial instrument, financial product or service, or to be used in any way for evaluating the merits of participating in any transaction, and shall not constitute a solicitation under any jurisdiction or to any person, if such solicitation under such jurisdiction or to such person would be unlawful.
\end{spacing}
}

\bibliography{bib}
\bibliographystyle{ACM-Reference-Format}

\end{document}

%% file: preamble.tex
\usepackage{amsfonts}
\usepackage{amsmath}
\usepackage{amsthm}

\usepackage{savesym}
\usepackage{bm}
\usepackage{algorithmic}
\usepackage{algorithm}

\newcommand{\inputListIndent}{\hspace{14.5pt}}

\savesymbol{comment}
\usepackage{easyReview}

\usepackage{comment}  

\usepackage{color}

\makeatletter
\newcommand\newtag[2]{#1\def\@currentlabel{#1}\label{#2}}
\makeatother

\def\BibTeX{{\rm B\kern-.05em{\sc i\kern-.025em b}\kern-.08emT\kern-.1667em\lower.7ex\hbox{E}\kern-.125emX}}

\newtheorem{definition}{Definition}[section]

\newcommand{\calA}{\mathcal{A}}

    \newcommand{\KeyGen}{\mathsf{KeyGen}}
    \newcommand{\Encrypt}{\mathsf{Encrypt}}
    
    \newcommand{\Decrypt}{\mathsf{Decrypt}}
    
    \newcommand{\sk}{\mathsf{sk}}
    \newcommand{\pk}{\mathsf{pk}}
    \newcommand{\evk}{\mathsf{evk}}
    \newcommand{\ct}{\mathsf{ct}}
    
    \newcommand{\Eval}{\mathsf{Eval}}

    \newcounter{protocol}
    
    
    \newcommand{\calD}{\mathcal{D}}
    \newcommand{\calS}{\mathcal{S}}
    \newcommand{\calU}{\mathcal{U}}
    \newcommand{\calP}{\mathcal{P}}
    \newcommand{\calM}{\mathcal{M}}
    
    \newcommand{\calT}{\mathcal{T}}

    \newcommand{\MLteam}{\textsf{ML} \hspace{1pt}}
    \newcommand{\MLOneRound}{\textsf{MLOneRound} \hspace{1pt}}
    \newcommand{\Compteam}{\textsf{Comp} \hspace{1pt}}
    \newcommand{\CompOneRound}{\textsf{CompOneRound} \hspace{1pt}}

    \newcommand{\waldtest}{\textsf{WaldTest} \hspace{1pt}}
    
    \newcommand{\select}{\textsf{Select} \hspace{1pt}}
    
    \newcommand{\Learn}{\textsf{Learn} \hspace{1pt}}
    
    \newcommand{\FHE}{\textsf{FHE} \hspace{1pt}}

    \DeclareMathOperator\se{se}
    
    \newcommand{\Adult}{\textsf{Adult} \hspace{1pt}}
    \newcommand{\Age}{\textsf{Age} \hspace{1pt}}